\documentclass{INTERSPEECH2023}

\usepackage{multicol}
\usepackage{multirow}
\usepackage{cite}
\usepackage{bbding}
\usepackage{hyperref}
\interspeechcameraready

\title{Knowledge Transfer from Pre-trained Language Models to Cif-based Speech Recognizers via Hierarchical Distillation}
\name{
    Minglun Han$^{1,2,\dagger}$, 
    Feilong Chen$^{1,3}$, 
    Jing Shi$^{1,\ddagger}$,
    Shuang Xu$^{1}$,
    Bo Xu$^{1,2,3}$
    \thanks{
        $\dagger$ This work was supported by the National Key Research and Development Program of China (2018AAA0100400), the Strategic Priority Research Program of the Chinese Academy of Sciences (No. XDA27030300) and the National Natural Science Foundation of China (62206294). $\ddagger$ Corresponding author.
    }
}
\address{
    $^1$Institute of Automation, Chinese Academy of Sciences\\
    $^2$School of Artificial Intelligence, University of Chinese Academy of Sciences\\
    $^3$School of Future Technology, University of Chinese Academy of Sciences
}
\email{
    \{hanminglun2018,chenfeilong2018,shijing2014\}@ia.ac.cn
}

\begin{document}

\maketitle

\begin{abstract}
Large-scale pre-trained language models (PLMs) have shown great potential in natural language processing tasks. Leveraging the capabilities of PLMs to enhance automatic speech recognition (ASR) systems has also emerged as a promising research direction. However, previous works may be limited by the inflexible structures of PLMs and the insufficient utilization of PLMs. To alleviate these problems, we propose the hierarchical knowledge distillation (HKD) on the continuous integrate-and-fire (CIF) based ASR models. To transfer knowledge from PLMs to the ASR models, HKD employs cross-modal knowledge distillation with contrastive loss at the acoustic level and knowledge distillation with regression loss at the linguistic level. Compared with the original CIF-based model, our method achieves 15\% and 9\% relative error rate reduction on the AISHELL-1 and LibriSpeech datasets, respectively.
\end{abstract}

\section{Introduction}

End-to-end (E2E) models have recently made remarkable progress on automatic speech recognition (ASR) tasks. Compared with hybrid models, E2E models are optimized in a unified structure. However, the tight integration in this unified structure hinders the infusion of linguistic knowledge and limits the use of large-scale textual corpora.

Currently, there are two popular approaches widely used to leverage unpaired text for E2E ASR models: language model (LM) fusion~\cite{gulcehre2015using,sriram2017cold,toshniwal2018comparison,shan2019component} and re-scoring~\cite{chan2016listen}. Apart from them, utilizing large-scale pre-trained language models (PLMs) to improve language modeling of ASR models~\cite{futami2020distilling,huang2021speech,chen2023xllm} is also a practical approach to make use of unpaired text dataset. PLMs possess powerful language modeling abilities, and their outputs contain rich linguistic information that can improve ASR language modeling~\cite{futami2020distilling,kubo2022knowledge}. Therefore, employing PLMs to improve speech recognition has gradually become an important research direction. Until now, the methods used to improve ASR with PLMs can be categorized into three classes: re-scorer based method, model-based method, and knowledge distillation based method. The re-scorer based methods~\cite{shin2019effective,Salazar2020MaskedLM,Chiu2021InnovativeBR,Futami2021ASRRA,Xu2022RescoreBERTDS} convert PLMs into re-scorers and use them to re-score the $N$-best lists or lattices from the first-pass decoding, while not changing the ASR model. Unlike the re-scorer based method, the model-based method and KD-based method focus on improving the ASR model itself. The model-based method refers to using PLM as part of the ASR model. For example, Huang et al.~\cite{huang2021speech} fine-tune PLM as an ASR model with acoustics as cues. Yi et al.~\cite{yi2021efficiently} use the CIF mechanism~\cite{dong2020cif} to combine pre-trained acoustic and language models in a unified structure. Following~\cite{yi2021efficiently}, Zheng et al.~\cite{zheng2021wav} and Deng et al.~\cite{deng2022model} integrate pre-trained acoustic and language models for low-resource ASR and non-autoregressive (NAR) ASR, respectively. However, directly deploying model-based methods may be challenging due to the large size and different structures of PLMs. The KD-based methods transfer knowledge from PLMs to ASR models via knowledge distillation~\cite{Hinton2015DistillingTK}. Futami et al.~\cite{futami2020distilling} distill knowledge from the BERT output distribution to the output distribution of the ASR model. Unlike the probability-based KD, the representation-based KD, which optimizes the similarity between teacher and student representations, transfers knowledge from PLMs to NAR ASR models~\cite{bai2021fast}. Furthermore, the representation-based KD is applied to various ASR models~\cite{kubo2022knowledge,deng2022distill}. However, most KD-based methods transfer the knowledge to only one of acoustics or linguistics and thus cannot fully leverage PLMs.

In this paper, to explore effective schemes of using PLMs in ASR, we propose a knowledge transfer strategy called hierarchical knowledge distillation (HKD). HKD transfers linguistic knowledge from PLMs to different levels of the ASR model, including the acoustic level. However, it is not easy to directly transfer linguistic knowledge to the acoustic level of E2E models. Unlike other E2E schemes, the continuous integrate-and-fire mechanism (CIF)\cite{dong2020cif}, which generates token-level acoustic representations aligned with the text, provides a natural option for the KD at the acoustic level. Thus, we develop the HKD based on the CIF-based ASR model. Inspired by contrastive knowledge distillation (CKD)\cite{tian2019contrastive,fu2021lrc}, we leverage contrastive loss to transfer the knowledge to the high-level acoustics of CIF-based ASR models. By pushing positive pairs together and negative pairs apart, the contrastive loss encourages the model to capture semantic alignment, giving CKD an advantage over losses that optimize similarity when distilling knowledge across different modalities and structures. At the linguistic level, we apply regression loss to transfer knowledge from the PLM to the linguistic representations. Unlike model-based methods, HKD does not require adapting the ASR model for PLMs. Compared with other representation-based KD methods, HKD transfers the knowledge into the ASR model at multiple levels and applies contrastive distillation to effectively bridge the semantic gap between acoustics and linguistics. Experiments show that HKD achieves 15\% and 9\% relative error rate reduction over the original CIF-based model on AISHELL-1 and LibriSpeech, respectively. The implementation is available on GitHub\footnote{\url{https://github.com/MingLunHan/CIF-HieraDist}}.

\section{Proposed method}

\begin{figure}[t]
  \centering
  \includegraphics[width=0.95\linewidth]{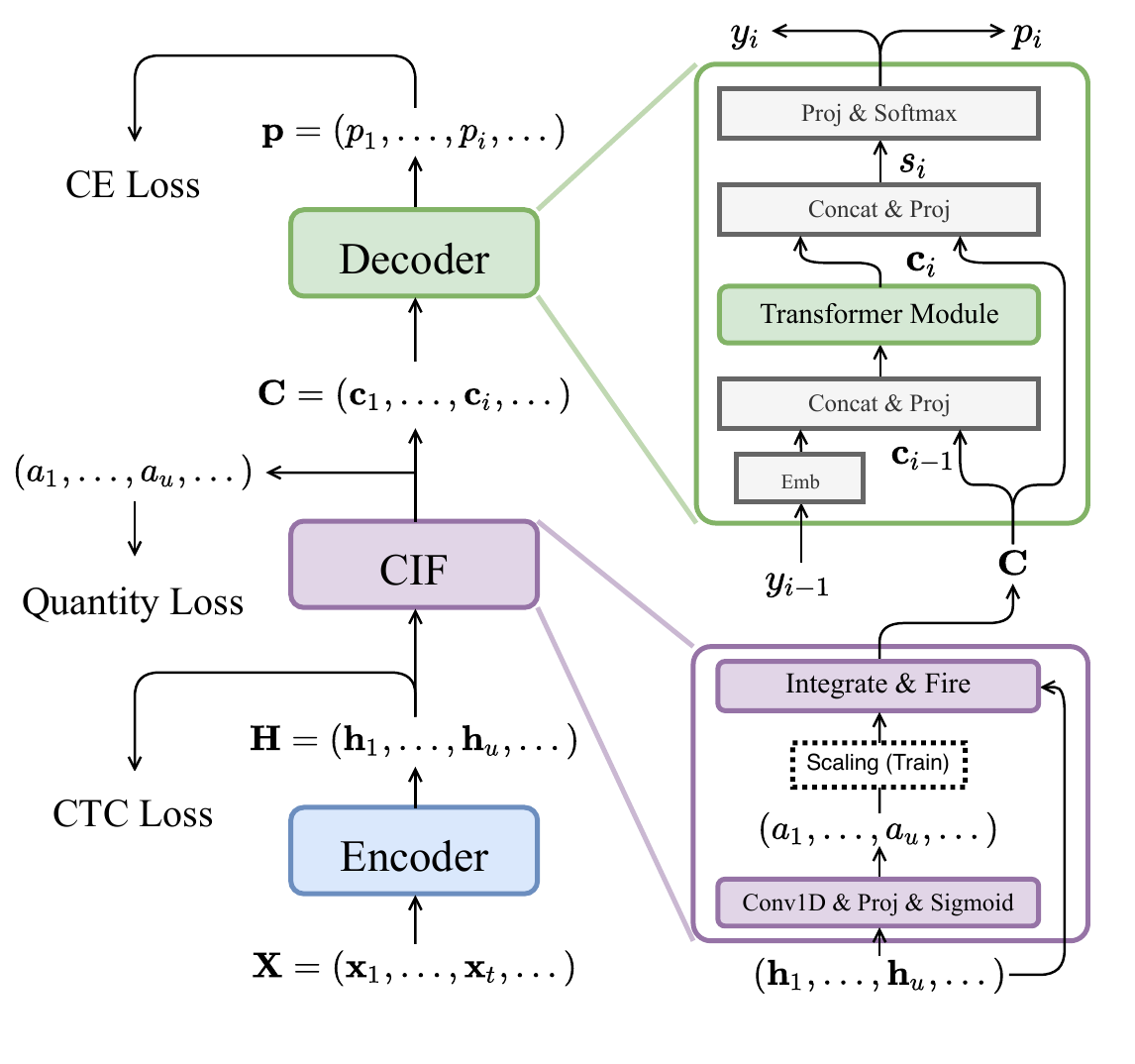}
    \vspace{-10pt}
  \caption{The CIF-based ASR model.}
  \label{fig_1}
  \vspace{-13pt}
\end{figure}

\subsection{Preliminaries}
\label{ssec:pre}

\subsubsection{Continuous Integrate-and-Fire based ASR model}
\label{sssec:cif}

Continuous Integrate-and-Fire (CIF)~\cite{dong2020cif}, a soft monotonic alignment mechanism, has been successfully applied to various ASR tasks~\cite{han2021coldec,han2022finecos}. As shown in Figure~\ref{fig_1}, the CIF-based ASR model in this work consists of an acoustic encoder, a CIF module, and a decoder. The acoustic encoder has a convolution front-end, and a conformer~\cite{gulati2020conformer} module. The CIF module has a 1-dimensional convolution layer and a fully-connected (FC) layer. The decoder, composed of FC layers and a transformer~\cite{vaswani2017attention} module, is an autoregressive decoder. 

The input feature sequence $\mathbf{X}=(\mathbf{x}_1,...,\mathbf{x}_t,...,\mathbf{x}_T)$ is first fed to the convolution front-end of the encoder. Then, the conformer module takes the outputs of the convolution front-end as inputs and outputs low-level acoustic sequence $\mathbf{H}=(\mathbf{h}_1,...,\mathbf{h}_u,...,\mathbf{h}_U)$. Note that the convolution front-end down-samples the inputs by 2, and the conformer module down-samples the inputs by 4 with two max-pooling layers. Next, $\mathbf{H}$ is delivered to the CIF module. In the CIF module, $\mathbf{H}$ are first passed through the 1-dimensional convolution layer, and then one FC layer with one output unit and a followed sigmoid activation is used to generate weights $\mathbf{a}=(a_1,...,a_u, ...,a_U)$ from outputs of the convolution layer. After that, The CIF module accumulates the weight $a_u$ along the time axis. When the accumulated weight exceeds a threshold $\beta$, a firing representing the acoustic boundary between adjacent tokens occurs. The weight of the firing time-step will be split into two parts: 1) the first part is used for the weight accumulation of the token before the boundary to make its accumulated weight reach $\beta$; 2) the second part is left for the accumulation of the token after the boundary. Further, the CIF module summarizes $h_u$ between adjacent acoustic boundaries via weighted sum with generated weights as weighting factors, and outputs high-level acoustic sequence $\mathbf{{C}}=({\mathbf{c}}_1, ..., {\mathbf{c}}_i, ..., {\mathbf{c}}_I)$. Finally, the decoder takes the high-level acoustic sequence $\mathbf{{C}}=({\mathbf{c}}_1, ..., {\mathbf{c}}_i, ..., {\mathbf{c}}_I)$ as inputs, and gives the final linguistic sequence $\mathbf{S}=(\mathbf{s}_1, ..., \mathbf{s}_i, ..., \mathbf{s}_I)$. 

\subsubsection{Large-scale pre-trained language models}
\label{sssec:plm}

\begin{figure}[t]
  \centering
  \includegraphics[width=0.95\linewidth]{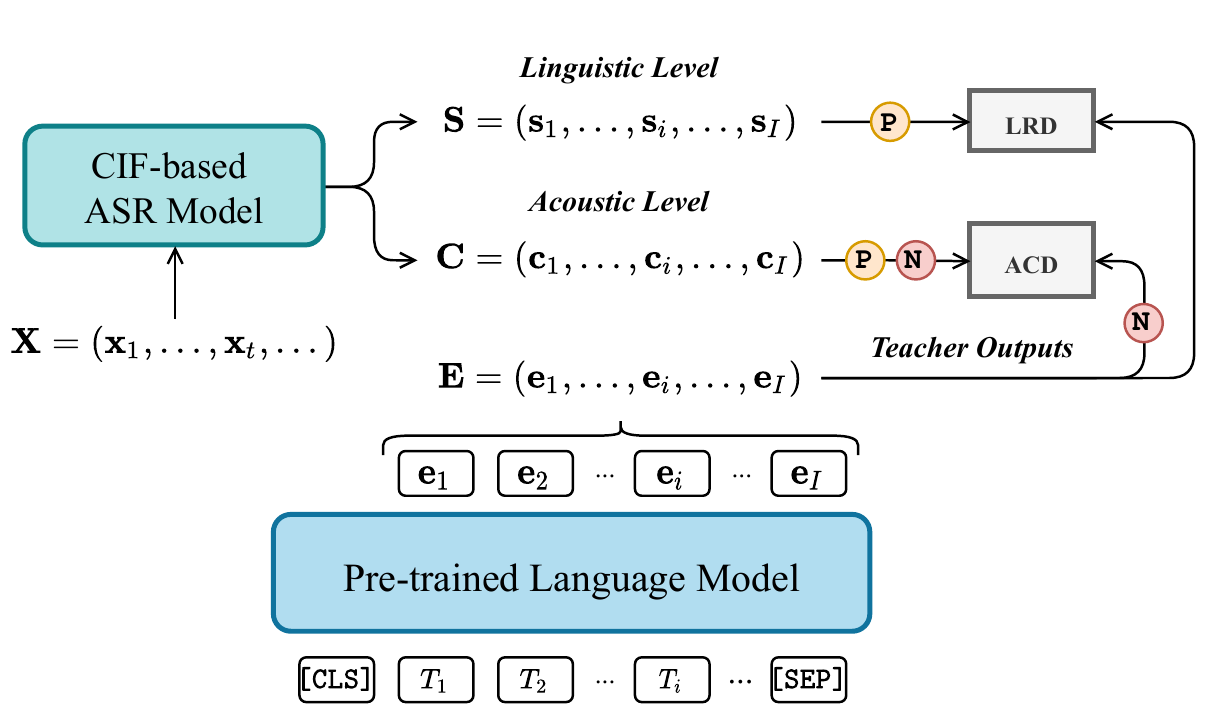}
  \caption{Hierarchical knowledge distillation. LRD denotes linguistic regression distillation, and ACD denotes acoustic contrastive distillation. \texttt{P} denotes projection, and \texttt{N} denotes L2 normalization.}
  \label{fig_2}
    \vspace{-10pt}
\end{figure}

PLMs trained on large-scale datasets, such as BERT~\cite{devlin2018bert} and GPT-2~\cite{radford2019language}, have been widely used in natural language processing tasks. PLMs possess strong modeling power and contain rich linguistic information, which is helpful to compensate for the language modeling of the ASR model. This work focuses on transferring knowledge from BERT-like PLM teachers to ASR students via knowledge distillation. Given text sequence $(T_1, ..., T_i, ..., T_{I-1}, \texttt{<EOS>})$ with length $I$, the input for PLMs is $(\texttt{[CLS]}, T_1, ..., T_i, ..., T_{I-1}, \texttt{[SEP]})$ with length $(I+1)$. As shown in Figure~\ref{fig_2}, to keep the strict alignment between the student ASR outputs and teacher PLM outputs, we ignore the PLM output corresponding to \texttt{[CLS]}. The final output sequence of teacher PLM is denoted as $\mathbf{E}=(\mathbf{e}_1, \mathbf{e}_2, ..., \mathbf{e}_i, ..., \mathbf{e}_I)$.

\subsection{Hierarchical knowledge distillation}
\label{ssec:hkd}

We propose hierarchical knowledge distillation (HKD) that transfers the knowledge from PLM to the CIF-based ASR model, as shown in Figure~\ref{fig_2}. ``Hierarchical" 1) describes the bottom-up ASR hierarchy: speech input is first transformed to low-level acoustic features $\mathbf{H}$, and then transformed to high-level acoustic features $\mathbf{C}$, and finally transformed to linguistic representations $\mathbf{S}$, and 2) describes the behavior of distillations that simultaneously happen at the acoustic level $\mathbf{C}$ and higher linguistic level $\mathbf{S}$. Such hierarchical distillation might better utilize the PLMs to enhance different aspects of ASR. The total loss of the ASR model with HKD is the sum of 1) ASR loss and 2) multi-level distillation losses. The total loss is written as
\begin{equation}
    \mathcal{L}_{Total} = \mathcal{L}_{ASR} + \lambda_{AD} \cdot \mathcal{L}_{AD} + \lambda_{LD} \cdot \mathcal{L}_{LD}
\end{equation}
where $\mathcal{L}_{ASR}$ is the ASR loss of the CIF-based model~\cite{dong2020cif}. $\mathcal{L}_{AD}$ and $\mathcal{L}_{LD}$ are acoustic distillation (AD) loss and linguistic distillation (LD) loss, respectively. $\lambda$ denotes the loss weight.

\subsubsection{Acoustic contrastive distillation}
\label{sssec:acd}

Considering that the CIF acoustic sequence $\mathbf{C}$ is strictly aligned with the text sequence during training~\cite{dong2020cif}, we can transfer the knowledge from PLMs to these high-level acoustic representations. However, there are two potential obstacles in this distillation process: 1) modal gap: although the CIF output $\mathbf{C}$ is aligned with text sequence, it is still closer to the acoustics (without linguistic contextual modeling); 2) structure gap: the acoustic encoder of the CIF-based model, which uses conformer structure and a weight accumulation mechanism, usually differs from the transformer structures of PLMs. Inspired by contrastive distillation~\cite{fu2021lrc}, we use contrastive loss for knowledge distillation across different modalities and structures. Compared with distillation losses that directly optimize the similarity metrics, contrastive loss forces the model to pull together the positive pairs and push apart the negative pairs. Thus, the model can capture the high-level semantic alignment between student and teacher, and better model semantics. More specifically, we use contrastive loss (based on InfoNCE~\cite{oord2018representation}) as the objective function for acoustic contrastive distillation (ACD). We project original student outputs in $\mathbf{C}$ to match the dimension of teacher output representation, and then normalize them. We denote the projected student outputs, final student outputs, and final teacher outputs as $\mathbf{\hat{C}}=(\hat{\mathbf{c}}_1, \hat{\mathbf{c}}_2, ..., \hat{\mathbf{c}}_i, ..., \hat{\mathbf{c}}_I)$, $\mathbf{\bar{C}}=(\bar{\mathbf{c}}_1, \bar{\mathbf{c}}_2, ..., \bar{\mathbf{c}}_i, ..., \bar{\mathbf{c}}_I)$ and $\mathbf{\bar{E}}=(\bar{\mathbf{e}}_1, \bar{\mathbf{e}}_2, ..., \bar{\mathbf{e}}_i, ..., \bar{\mathbf{e}}_I)$, respectively. The contrastive loss is defined as
\begin{equation}
\mathcal{L}^{cont}_{AD} = -\frac{1}{N}\sum_{n=1}^{N}\frac{1}{I^n}\sum_{i=1}^{I^n}log \frac{s(\mathbf{\bar{c}}_{i},\mathbf{\bar{e}}_{i})}{\sum_{k=1}^{K}{s(\mathbf{\bar{c}}_i, \mathbf{\bar{e}}_{n,i,k}^{-})+s(\mathbf{\bar{c}}_{i}, \mathbf{\bar{e}}_{i})}} , 
\end{equation}
where $s(\mathbf{x},\mathbf{y})$ is equal to $exp(\langle{\mathbf{x},\mathbf{y}}\rangle/\tau)$, and $\langle{\mathbf{x},\mathbf{y}}\rangle$ denotes the inner-product of $\mathbf{x}$ and $\mathbf{y}$. $N$ and $I^{n}$ denote the batch size and the text length of the $n$-th audio sample, respectively. $\tau$ and $K$ denote the temperature and the number of negative samples for contrastive loss. $\bar{\mathbf{c}}_i$ represent the $i$-th student token query of the $n$-th sample. $\bar{\mathbf{e}}_i$ represents the positive teacher token representation that matches $\bar{\mathbf{c}}_i$. $\bar{\mathbf{e}}_{n,i,k}^{-}$ represents the $k$-th negative teacher token representation sampled from all teacher token representations (except the positive one) of the current batch.

Apart from the contrastive loss, we also try to conduct distillation with the mean square error (MSE) loss or the cosine embedding (COS) loss for comparison. They can be written as 
\begin{equation}
\mathcal{L}^{mse}_{AD} = \alpha_{mse} \cdot \frac{1}{N}\sum_{n=1}^{N}\frac{1}{I^{n}}\sum_{i=1}^{I^{n}}\sum_{d=1}^{D}(\hat{c}_{i,d}^n - e_{i,d}^n)^2 ,
\end{equation}
\begin{equation}
\mathcal{L}^{cos}_{AD} = \alpha_{cos} \cdot \frac{1}{N}\sum_{n=1}^{N}\frac{1}{I^{n}}\sum_{i=1}^{I^{n}}{(1 - cosine(\mathbf{\hat{c}_i}, \mathbf{e_i}))} ,
\end{equation}
where $D$ is the dimension of teacher representations. Coefficients $\alpha_{mse}$ and $\alpha_{cos}$ scale losses to achieve the balance.

\subsubsection{Linguistic regression distillation}
\label{ssec:lrd}

We use regression loss to distill the knowledge from PLMs to the final linguistic representations of the CIF-based model. Using regression loss to transfer the knowledge to ASR models has been proven effective~\cite{bai2021fast}. However, it is still uncertain whether this method works for the CIF-based ASR models. Specifically, we use MSE loss as the objective function for linguistic regression distillation (LRD). Given the projected final state of the decoder $\mathbf{\hat{\mathbf{S}}}=(\hat{\mathbf{s}}_1, \hat{\mathbf{s}}_2, ..., \hat{\mathbf{s}}_i, ..., \hat{\mathbf{s}}_I)$ as student outputs and the PLM outputs $\mathbf{E}$ as teacher outputs, MSE loss can be defined as
\begin{equation}
\mathcal{L}^{mse}_{LD} = \alpha_{mse} \cdot \frac{1}{N}\sum_{n=1}^{N}\frac{1}{I^{n}}\sum_{i=1}^{I^{n}}\sum_{d=1}^{D}(\hat{s}_{i,d}^n - e_{i,d}^n)^2 .
\end{equation}

\section{Experimental setup}
\label{sec:experimental_settings}

\subsection{Datasets and metrics}
\label{ssec:DandM}

We evaluate our method on a Mandarin Chinese dataset AISHELL-1~\cite{bu2017aishell} and an English dataset LibriSpeech~\cite{Panayotov2015LibrispeechAA}. We extract 80-channel filterbank features computed from a 25ms window with a stride of 10ms. For AISHELL-1, the output vocabulary contains 4230 characters and four special tokens \texttt{<PAD>}, \texttt{<EOS>}, \texttt{<BOS>}, \texttt{<UNK>}. For LibriSpeech, because PLM and the English ASR model use different output vocabularies, we directly use the vocabulary of PLM for the ASR model for the convenience of distillation. We use the character error rate (CER) and word error rate (WER) to measure ASR performance for Chinese and English, respectively.

\subsection{Configurations}
\label{ssec:config}

For Chinese, the encoder of the ASR model consists of a convolution front-end and a conformer module. The convolution front-end is a 2-dimensional convolution layer with output channels $128$, kernel size $3$, and strides $2$. The conformer module consists of 15 conformer blocks with $d_{model}=256$, $d_{ffn}=2048$ and $h=4$, kernel size $15$ (for depth-wise convolution), and 2 max-pooling layers after the 5th and the 10th blocks. The CIF module contains a 1-dimensional convolution layer with output channels $256$, kernel size $3$ and strides $1$, and an FC layer followed by the sigmoid activation. The decoder consists of several FC layers and a transformer module, which consists of $2$ transformer blocks with $d_{model}=256$, $d_{ffn}=2048$, and $h=4$. For English, the hidden size and the number of attention heads are set to 512 and 8, respectively. The number of output channels of the convolution layer in the CIF module is 512.

During training, we apply dropout for conformer blocks (0.1), transformer blocks (0.2), and the convolution layer (0.2) in the CIF module. In addition, we apply SpecAugment~\cite{park2019specaugment} with $F=27$, $m_F=2$, $T=50$, $m_T=2$ and $p=1.0$. We apply label smoothing with $\epsilon=0.1$. We train the models with the Adam optimizer~\cite{kingma2014adam} with $\beta_1 = 0.9$, $\beta_2 = 0.98$, $lr=3\text{e-}{4}$ and a weight decay of $0.01$. The weights of cross-entropy loss, connectionist temporal classification loss, and quantity loss are set to $1.0$, $0.5$ and $1.0$, respectively. The threshold $\beta$ of the CIF mechanism is $1.0$. The scaling strategy and tail handling in~\cite{dong2020cif} are applied. The weights of distillation losses are tuned on the dev set and chosen from $\{0.01, 0.1, 0.2, 0.5, 1.0\}$. $\alpha_{mse}$ and $\alpha_{cos}$ are set to $0.01$ and $10$, respectively. The PLMs used for distillation are {bert-base-chinese}\footnote{\url{https://huggingface.co/bert-base-chinese}} for Chinese and {bert-base-uncased}\footnote{\url{https://huggingface.co/bert-base-uncased}} for English. Note that all PLMs are fixed during training. During inference, we use beam search with beam size 10. For Chinese, we use a 16-layer Transformer LM (trained with the text of all training data) via shallow fusion~\cite{gulcehre2015using} with the interpolation weight tuned on the dev set.

\section{Results}
\label{sec:results}

\subsection{Results on AISHELL-1}

\begin{table}[t]
    \centering
    \caption{Main results on AISHELL-1 (CER \%).}
    \vspace{-5pt}
    \resizebox{\linewidth}{!}{
    \begin{tabular}{l|cc|cc}
    \toprule
    \textbf{Model} & \textbf{LM} & \textbf{\# Param} & \textbf{dev (\%)} & \textbf{test (\%)} \\
    \midrule
    ESPnet Conformer~\cite{watanabe2018espnet} & \XSolidBrush & 46 M & 4.5 & 4.9 \\
    ESPnet Conformer~\cite{watanabe2018espnet} & \Checkmark & 46 M & 4.4 & 4.7 \\
    Branchformer~\cite{peng2022branchformer} & \XSolidBrush & 45 M & 4.2 & 4.4 \\
    WeNet~\cite{yao2021wenet} & \Checkmark & 46 M & - & 4.4 \\
    Icefall & \Checkmark & - & - & 4.3 \\ 
    Neural Transducer~\cite{tian2022integrating} & \Checkmark & 90 M & 3.8 & 4.1 \\
    \midrule
    CIF & \XSolidBrush & 47 M & 4.5 & 4.9 \\
    \quad + ACD & \XSolidBrush & 47 M & 4.2 & 4.7 \\
    \quad + LRD & \XSolidBrush & 47 M & 4.0 & 4.5 \\
    \quad + HKD & \XSolidBrush & 47 M & 3.8 & 4.2 \\
    \midrule
    CIF & \Checkmark  & 47 M & 4.4 & 4.8 \\
    \quad + ACD & \Checkmark & 47 M & 4.2 & 4.6 \\
    \quad + LRD & \Checkmark & 47 M & 4.0 & 4.4 \\
    \quad + HKD & \Checkmark & 47 M & \textbf{3.8} & \textbf{4.1} \\
    \bottomrule
    \end{tabular}%
    }
    \label{table_1}%
    \vspace{-5pt}
\end{table}%

The experiments are conducted on the AISHELL-1 dataset. As depicted in Table~\ref{table_1}, we first compare the CIF-based ASR model with models in other literature. With comparable model parameters, the CIF-based ASR model achieves comparable performance to the ESPnet conformer~\cite{watanabe2018espnet}, with or without LM. Then, we use the CIF-based ASR model with LM as the baseline to evaluate the effectiveness of our method. We experiment with three settings: ACD only, LRD only, and HKD which combines ACD and LRD. The results reported in the last three rows show that ACD, LRD, and HKD achieve about 4\%, 8\%, and 15\% relative error rate reduction, respectively. With the help of PLMs, the CIF-based model achieves comparable performance with the strong baseline~\cite{tian2022integrating}. We can conclude that 1) both ACD and LRD can improve ASR performance; 2) HKD could further enhance the ASR model, which proves the complementary nature of LRD and ACD. Note that our method brings no additional inference cost.

\begin{table}[t]
    \centering
    \caption{Comparison between contrastive loss and other distillation losses (CER \%). AD represents acoustic distillation. MSE, COS, and CONT represent mean square error loss, cosine embedding loss, and contrastive loss, respectively.}
    \vspace{-5pt}
    \resizebox{\linewidth}{!}
    {
    \begin{tabular}{lccccc}
        \toprule 
        \multirow{2}[4]{*}{\textbf{Model}} & \multirow{2}[4]{*}{\textbf{LRD}} & \multirow{2}[4]{*}{\textbf{AD}} & \multicolumn{1}{c}{\multirow{2}[4]{*}{\textbf{AD Loss}}} & \textbf{w/o LM} & \textbf{w/ LM}  \\
        \cmidrule{5-6} & & & & \textbf{dev / test} & \textbf{dev / test} \\
        \cmidrule{1-6} \multirow{8}[4]{*}{CIF} 
            & \XSolidBrush & \XSolidBrush & -   & 4.5 / 4.9 & 4.4 / 4.8 \\
            & \XSolidBrush & \Checkmark & MSE   & 4.4 / 4.9 & 4.4 / 4.8 \\
            & \XSolidBrush & \Checkmark & COS   & 4.5 / 4.9 & 4.4 / 4.8 \\
            & \XSolidBrush & \Checkmark & CONT  & 4.2 / 4.7 & 4.2 / 4.6 \\
            \cmidrule{2-6}          
            & \Checkmark & \XSolidBrush & -   & 4.0 / 4.5 & 4.0 / 4.4 \\
            & \Checkmark & \Checkmark & MSE   & 4.0 / 4.5 & 4.0 / 4.5 \\
            & \Checkmark & \Checkmark & COS   & 4.1 / 4.5 & 4.0 / 4.4 \\
            & \Checkmark & \Checkmark & CONT  & \textbf{3.8 / 4.2} & \textbf{3.8 / 4.1} \\
        \bottomrule
    \end{tabular}%
    }
    \label{table_2}%
    \vspace{-10pt}
\end{table}

We compare the contrastive loss with other losses that optimize the similarity metrics directly. We conduct experiments under two settings: a CIF-based ASR baseline and a CIF-based ASR baseline with LRD. As shown in Table~\ref{table_2}, the contrastive loss outperforms MSE loss and COS loss. This result may result from the fact that contrastive loss encourages the model to learn semantic alignments rather than strictly optimize the similarity metrics. Thus, the contrastive loss can perform better under the cross-modal distillation settings. The weight of MSE loss is set to 1.0. The weight of COS loss is set to 0.2. The weight of contrastive loss, the temperature $\tau$, and the number of negative samples $K$ are set to 1.0, 0.02, and 700, respectively.

\begin{figure}[t]
    \centering
    \includegraphics[scale=0.43]{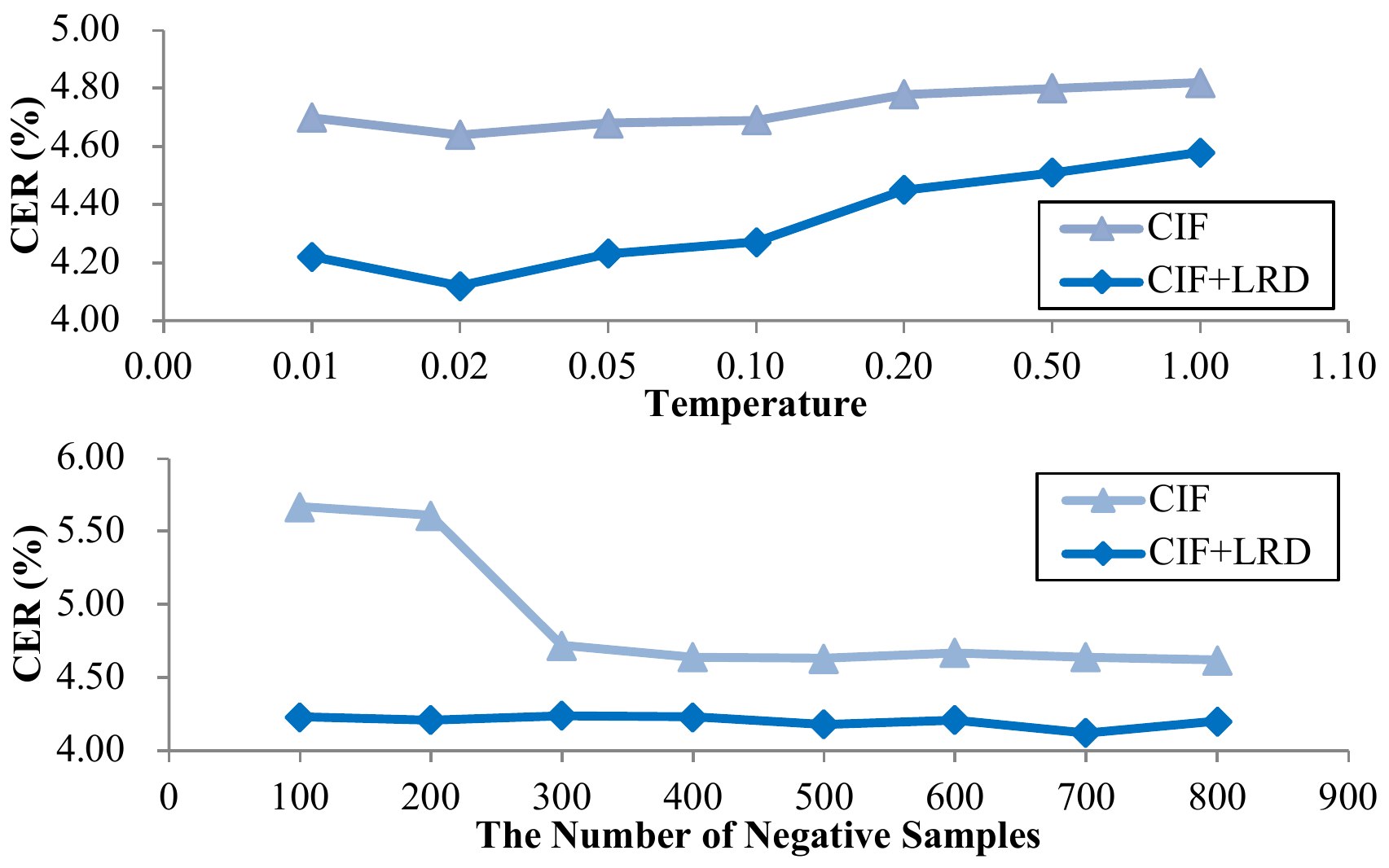}
    \vspace{-5pt}
    \caption{Effects of the temperature and the number of negative samples.}
    \label{fig_3}
    \vspace{-10pt}
\end{figure}

We explore the effects of $\tau$ and $K$ on ACD with a CIF-based ASR model and a CIF-based ASR model with LRD. Figure~\ref{fig_3} shows the trend of CER as the temperature increases. Obviously, increasing $\tau$ leads to degradation. With $\tau$ chosen from $\{0.01, 0.02, 0.05\}$, ACD make CER fluctuate around 4.2\% and provide stable improvements. We set $\tau$ to 0.02 to report the best results. Figure~\ref{fig_3} shows the trend of CER as $K$ increases. Generally speaking, more negative samples will lead to better performance. We report the best results with $K=700$. When $K$ is chosen from $\{100, 200, 300, 400, 500, 600\}$, the ASR model with HKD almost achieves comparable performances (around 4.2\%). However, when we remove LRD, a severe deterioration occurs for settings with small $K$, which implies that LRD helps to stabilize the training of ACD. Since increasing $K$ leads to more training memory cost, it is necessary to choose a compromised $K$ to achieve comparable performance in practical usage.

\subsection{Results on LibriSpeech}

As shown in Table~\ref{table_librispeech}, our methods consistently improve performance on dev sets and test sets, which demonstrates the efficacy of our methods on the English dataset. Using ACD and LRD simultaneously, we can achieve a relative WER reduction of 9\% on both test-clean and test-other. Our methods yield a lower relative performance gain on the English dataset than on the Chinese dataset. We hypothesize that the difference in the property of output modeling units may result in this discrepancy. In contrast to the Chinese modeling units (characters), the English modeling units (especially some intra-word subwords) may lack clear acoustic boundaries. Therefore, it is difficult for English to learn a proper cross-modal alignment between acoustics and linguistics via contrastive knowledge distillation.

\begin{table}[t]
    \centering
    \caption{Main results on LibriSpeech (WER \%).}
    \vspace{-5pt}
    {
    \begin{tabular}{l|p{0.7cm}<{\centering}p{0.7cm}<{\centering}p{0.7cm}<{\centering}p{0.7cm}<{\centering}}
    \toprule
    \textbf{Model} & \textbf{dev clean} & \textbf{dev other} & \textbf{test clean} & \textbf{test other} \\
    \midrule
    CIF & 3.0 & 7.3 & 3.3 & 7.7 \\
    \quad + ACD & 3.0 & 7.2 & 3.2 & 7.3 \\
    \quad + LRD & 2.8 & 6.9 & 3.1 & 7.1 \\
    \quad + HKD & \textbf{2.7} & \textbf{6.9} & \textbf{3.0} & \textbf{7.0} \\
    \bottomrule
    \end{tabular}%
    }
    \label{table_librispeech}%
    \vspace{-12pt}
\end{table}%

\section{Conclusion}
\label{sec:conclusion}

In this work, we introduce a hierarchical knowledge distillation strategy to transfer PLM knowledge to different levels of the CIF-based ASR model. Specifically, we use acoustic contrastive distillation at the acoustic level and linguistic regression distillation at the linguistic level. Compared to the CIF-based ASR baseline, our method brings 15\% relative CER reduction on AISHELL-1 and 9\% relative WER reduction on LibriSpeech. We will explore our methods with larger-scale language models in the future.

\bibliographystyle{IEEEtran}
\bibliography{mybib}

\end{document}